\documentclass{bmvc2k}
\usepackage{multirow}

\title{Decoupling Forgery Semantics for Generalizable Deepfake Detection}

\addauthor{Wei Ye}{weiye@email.ncu.edu.cn}{1}
\addauthor{Xinan He}{shahur@email.ncu.edu.cn}{1}
\addauthor{Feng Ding*}{fengding@ncu.edu.cn}{1}

\addinstitution{
 Nanchang University\\ 
 Nanchang, China
}

\runninghead{YE \emph{et al.}}{Decoupling Forgery Semantics}

\def\eg{\emph{e.g}\bmvaOneDot}
\def\Eg{\emph{E.g}\bmvaOneDot}
\def\etal{\emph{et al}\bmvaOneDot}

\begin{document}

\maketitle

\begin{abstract}
In this paper, we propose a novel method for detecting DeepFakes, enhancing the generalization of detection through semantic decoupling. There are now multiple DeepFake forgery technologies that not only possess unique forgery semantics but may also share common forgery semantics. The unique forgery semantics and irrelevant content semantics may promote over-fitting and hamper generalization for DeepFake detectors. For our proposed method, after decoupling, the common forgery semantics could be extracted from DeepFakes, and subsequently be employed for developing the generalizability of DeepFake detectors. Also, to pursue additional generalizability, we designed an adaptive high-pass module and a two-stage training strategy to improve the independence of decoupled semantics. Evaluation on FF++, Celeb-DF, DFD, and DFDC datasets showcases our method's excellent detection and generalization performance. Code is available at: \url{https://github.com/leaffeall/DFS-GDD}.

\end{abstract}

\section{Introduction}
\label{sec:intro}
In recent years, the emergence of Deepfake technology has introduced significant challenges to the authenticity and trustworthiness of visual content online. These sophisticated algorithms can produce synthetic facial images and videos with remarkable realism, making it increasingly difficult to distinguish between genuine and fake media. Albeit the development of various deepfake detection methods \cite{Chen2021Attentive, Qian2020f3net, Zhou2023Exposing, Luo2021Generalizing, DBLP:conf/bmvc/BaldassarreDPW22, DBLP:conf/bmvc/HuangHCKLS21,DBLP:journals/access/GuoHWCL22, DBLP:journals/pr/PuHWLHZSSWL22}, their performance often suffers when applied to real-world scenarios due to a lack of robustness \cite{ding2022exs} and generalizability \cite{Lin2024Fairness}.

\begin{figure}[tb]
  \centering
  \includegraphics[height=6.5cm,width=\textwidth,keepaspectratio]{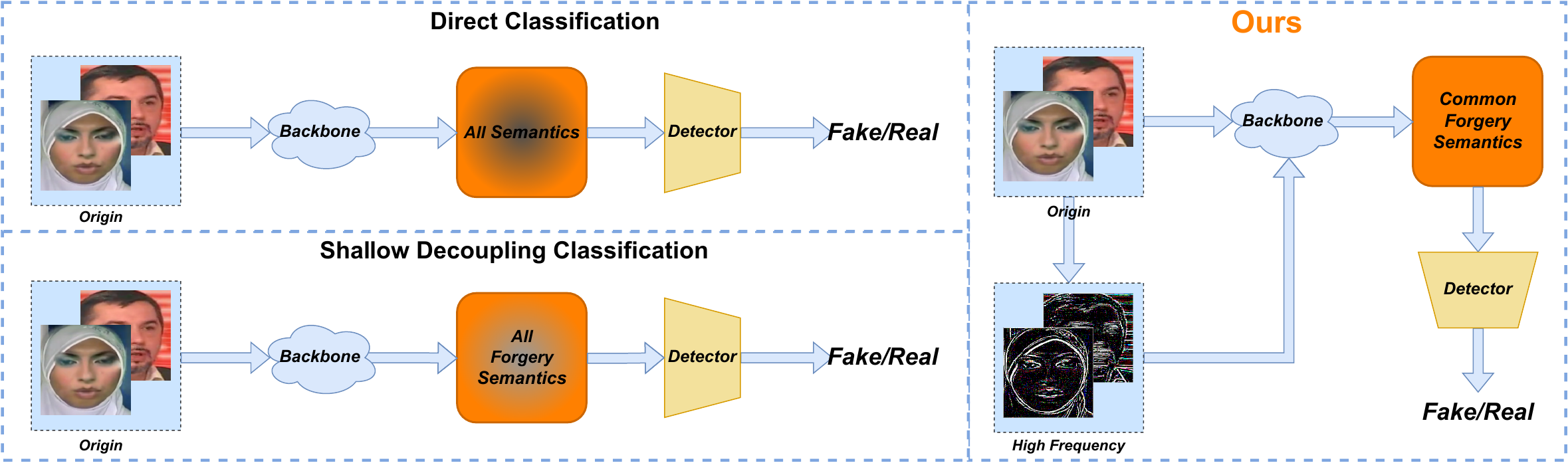}
  \caption{Comparison of Our Method with Existing Techniques. Orange represents semantics enhancing generalization, while grey indicates hindering generalization.
  }
  \label{figure:introduction}
\end{figure}

Existing methods typically fall into two categories \cite{Yan2023UCF}: direct classification and shallow decoupling classification. However, these approaches often suffer from interference by irrelevant content semantics or overlook common and unique forgery semantics, leading to limited generalization \cite{Lin2024Fairness}, as illustrated in Figure~\ref{figure:introduction}. They often perform well in intra-domain detection, i.e., with testing datasets similar to the training dataset, but experience severe performance degradation in cross-domain detection, i.e., with testing datasets different from the training dataset.

Also, relying solely on semantics extracted from images for training can lead to detectors overly relying on color textures specific to certain forgery methods \cite{Luo2021Generalizing}. Although some methods attempt to incorporate frequency domain data through techniques like Fourier transforms \cite{Chen2021Attentive} and Discrete Cosine Transform \cite{Zhou2023Exposing}, they often ignore the correlation between frequency domain and traditional color textures.

To address these challenges, we propose a method that leverages high-frequency features and deep decoupling to extract common forgery semantics, unique forgery semantics, and irrelevant content semantics, separating them as independent semantics for forensics purposes. Our approach consists of two training stages. In the first stage, high-frequency features along with images are utilized as inputs. We employ multi-scale high-frequency feature extraction and fusion modules, leveraging vision transformers for global feature extraction. In the second stage, we refine common forgery semantics to enhance generalizable detection performance.

Our contributions are as follows:

1. We propose a forensics model that leverages high-frequency features using our designed adaptive high-pass filter (AHF), combined with deep decoupling to extract common forgery semantics for generalizable DeepFake detection.

2. We introduce two modules: a multi-scale high-frequency feature extraction module (MHFE) and a multi-scale high-frequency feature fusion module (MHFF) to improve the independence and efficiency of extracted forgery semantics.


3. Other than the satisfying intra-domain detection performance, we demonstrate the superior generalization capability of our model in cross-domain deepfake detection scenarios through sufficient evaluations. Also, the impacts of the proposed modules are validated with ablation studies. 

The remainder of the paper is organized as follows. In the following section, we survey related works. The proposed method is described in Section 3 and the evaluation results are reported in Section 4. At last, we conclude the paper. 

\section{Related Work}
\textbf{Deepfake detection.} Deepfake detection methods are mainly divided into spatial-based forgery detection and frequency-based forgery detection. Spatial-based forgery detection primarily focuses on examining appearance features in the spatial domain \cite{DBLP:journals/access/GuoHWCL22, DBLP:journals/pr/PuHWLHZSSWL22, DBLP:conf/bmvc/HuangHCKLS21}. These methods often achieve satisfactory performance in intra-domain evaluation but encounter significant performance degradation in cross-domain testing. Some important features are difficult to be discovered when only utilizing RGB/spatial information, but these clues are often better revealed in the frequency domain \cite{song2022adaptive}. Therefore, some methods attempt to improve generalization by leveraging frequency domain components \cite{Chen2021Attentive, Qian2020f3net, Zhou2023Exposing, Luo2021Generalizing, DBLP:conf/bmvc/BaldassarreDPW22}. However, the improvement of these techniques in generalizing to unknown forgery technologies remains limited \cite{ding2021anti, ding2022securing, fan2024synthesizing,DBLP:conf/bmvc/GaoLXXC23}.
\\
\textbf{Decoupling in Deepfake Detection.} Decoupling, a technique breaking down complex semantics into simpler, more discriminative variables, has gained attention \cite{Liang2022Exploring}. Some studies have attempted to decouple forgery semantics for detection \cite{Hu2021Improving, FU2023104686, ba2024exposing}. Additionally, Liang \emph{et al.} \cite{Liang2022Exploring} reinforced feature independence through content consistency and global representation contrastive constraints. Recent research \cite{Yan2023UCF, Lin2024Fairness} further attempts to decompose manipulation-related semantics into unique and common forgery semantics, utilizing common forgery semantics for detection, alleviating the issue of performance degradation in cross-domain detection.

\section{Proposed Method}
\subsection{Motivation}
We aim to improve the generalization performance of deepfake detection by addressing three main challenges. Firstly, irrelevant content semantics may lead to the over-fitting of detectors, thereby hindering generalization. Secondly, the diversity of forgery techniques leads to distinct forgery artifacts, making detectors trained on specific sets of artifacts less effective at identifying unseen forgeries. Additionally, relying solely on RGB information may lead to over-reliance on specific color textures, while extracting features in a multi-modal manner can make the method more effective \cite{song2022adaptive, Zhou2023Exposing, Luo2021Generalizing}.
\begin{figure}[t]
  \centering
\includegraphics[width=\linewidth,keepaspectratio]{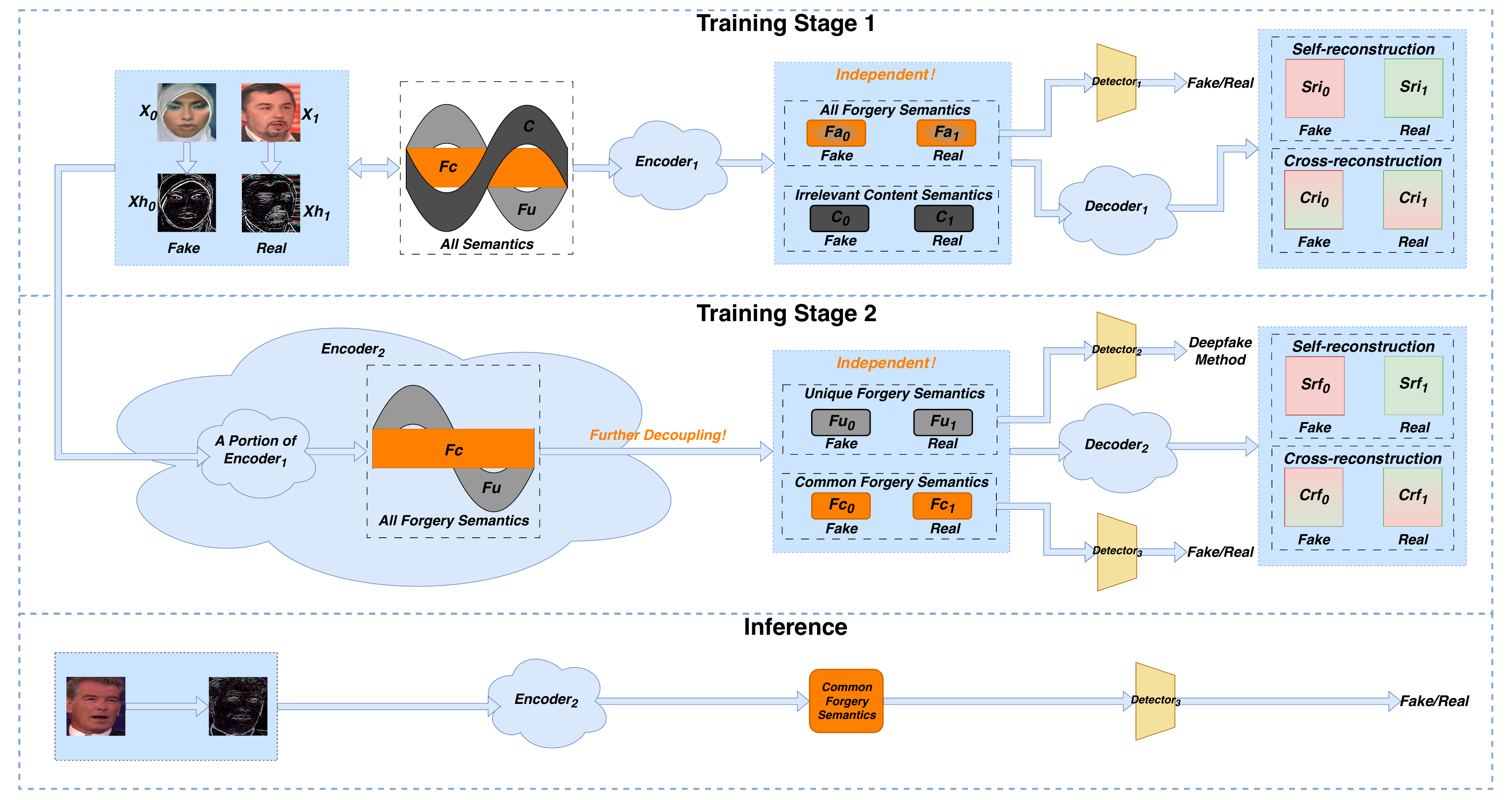}
\caption{The overview of our method. In the entangled semantics, dark gray represents irrelevant content semantics $C$, light gray represents unique forgery semantics $Fu$, and orange represents common forgery semantics $Fc$. $Encoder_2$ utilizes branches to extract all forgery semantics from $Encoder_1$. Both $Decoder_1$ and $Decoder_2$ include processes for self-reconstruction and cross-reconstruction.}
  \label{figure:overview}
\end{figure}
\subsection{Overview}
Our method involves two-stage training, outlined in Figure~\ref{figure:overview}. In training stage 1, a pair of real and fake images along with their corresponding high-frequency features are inputs. $Encoder_1$ decouples them into irrelevant content semantics and all forgery semantics $Fa$, which are then identified by $Detector_1$. This stage ensures feature disentanglement through self-reconstruction and cross-reconstruction using $Decoder_1$. $Fa$ is composed of common forgery semantics $Fc$ and unique forgery semantics $Fu$, intertwined as follows:
\begin{equation}
    Fa = \left[Fc, Fu\right] 
    \label{eq:1}
\end{equation}
In training stage 2, the identical real and fake image pairs are employed as inputs again. $Encoder_2$ extracts $Fa$ from $Encoder_1$, further disentangling them into $Fc$ and $Fu$. $Detector_2$ identifies $Fu$, while $Detector_3$ identifies $Fc$. This stage ensures semantics disentanglement through self-reconstruction and cross-reconstruction using $Decoder_2$. During inference, only $Fc$ from $Encoder_2$ are used for real-fake classification, followed by $Detector_3$.

\subsection{Training Stage 1}
\textbf{Encoder1.} SwiftFormer \cite{Shaker2023SwiftFormer} introduces an efficient additive attention mechanism to replace costly matrix operations, achieving a better balance between accuracy and efficiency. We use SwiftFormer-L1 to extract irrelevant content from images. Additionally, we employ Xception \cite{Chollet2017Xception} for RGB and high-frequency feature extraction. To improve the efficiency of high-frequency features, we design an adaptive high-pass filter (AHF), inspired by the Gaussian filter. The convolution kernel of AHF is devised using the following formula, overcoming the fixed parameter limitations in traditional high-pass filtering:
\begin{equation}
    g(x,y,\sigma) =  E - \frac{{\text{G}(x,y,\sigma)}}{{\sum_{x,y} \text{G}(x,y,\sigma)}} 
    \label{eq:2}
\end{equation}

where \( g \) represents the convolution kernel of AHF, \( x \) and \( y \) denote positions in the convolution kernel, \( \sigma \) is set to 1.0, \( E \) is a matrix with a center value of 1 and the rest as 0, \( G \) is the Gaussian distribution function.
\\
Furthermore, to ensure the high-pass filtering characteristics of AHF, after initialization and  each backpropagation, we reset the center element to -1 and normalize the remaining elements:
\begin{equation} 
    \begin{cases}
        \widehat{g}(x,y)=-1 & \text{if } (x,y) = (0,0) \\
        \widehat{g}(x,y)=\frac{g(x,y)}{\sum_{x,y}g(x,y) - g(0,0)} & \text{if } (x,y) \neq (0,0)
    \end{cases}
    \label{eq:3}
\end{equation}

where \( \widehat{g} \) represents the updated convolution kernel of AHF, \( g \) represents the original convolution kernel of AHF, and \( x \) and \( y \) denote positions in the convolution kernel, with \( (x, y) = (0, 0) \) indicating the center of the kernel.
\\
To enhance the effectiveness of semantics for forensics purposes, we design a multi-scale high-frequency feature extraction module (MHFE) based on the adaptive high-pass filter. We also employ the Pag feature fusion method \cite{Xu2023PIDNet} to design a multi-scale high-frequency feature fusion module (MHFF), which allows us to integrate high-frequency features into the RGB stream, preventing over-reliance on specific color textures. In Section 4, we conducted ablation studies to justify the effectiveness of the proposed modules. The supplementary material illustrates the network architecture of $Encoder_1$ and demonstrates the process for inputs $X_1$ and $Xh_1$. The entire process is represented as:
\begin{equation}
    Encoder_1(X_i, Xh_i) = C_i, Fa_i
    \label{eq:4}
\end{equation}

where $i = (0, 1)$ denotes the image index. $X_i$, $Xh_i$, $C_i$, and $Fa_i$ represent the RGB information, high-frequency features, irrelevant content semantics, and all forgery semantics, respectively.
\\
\textbf{Decoder1.} We designed a dual-channel network in $Decoder_1$. One channel simultaneously utilizes convolutional layers and SwiftFormer \cite{Shaker2023SwiftFormer} to process irrelevant content semantics, while the other channel solely adopts convolutional layers to handle common and unique forgery semantics. These semantics are fused to reconstruct the image, enhancing decoupling capability using self-reconstruction and cross-reconstruction techniques \cite{Yan2023UCF}. The supplementary material illustrates the network framework of $Decoder_1$ and demonstrates the process for inputs $C_1$ and $Fa_1$. The overall process is represented as:
\begin{equation}
    Decoder_1(C_{i_1}, {Fa}_{i_2}) = 
    \begin{cases} 
      {Sri}_{i_1} & \text{if } i_1 = i_2 \\
      {Cri}_{i_1} & \text{if } i_1 \neq i_2 
    \end{cases}
    \label{eq:5}
\end{equation}

where $i_1 = (0, 1)$ and $i_2 = (0, 1)$ denote the image indices. $C_{i_1}$, $Fa_{i_2}$, $Sri_{i_1}$, and $Cri_{i_1}$ represent irrelevant content semantics, all forgery semantics, self-reconstructed images, and cross-reconstructed images, respectively.
\\
\textbf{Objective Function.} The framework's overall loss function combines two distinct components using a weighted sum: forgery semantics detection loss and reconstruction loss.

\underline{Classification Loss.} For detecting all forgery semantics, we use cross-entropy loss:
\begin{equation}
    L_{cls} = L_{ce}(Detector_1(Fa_i), y_i)
    \label{eq:6}
\end{equation}

where $i = (0, 1)$ denotes the image indices. $L_{ce}$, $Fa_i$, and $y_i$ (fake, real) represent the cross-entropy loss, all forgery semantics, and binary classification labels, respectively.

\underline{Reconstruction Loss.} We employ an $L_1$ reconstruction loss to maintain feature completeness and image consistency:
\begin{equation}
    L_{rec} = \|X_{i_1}-Decoder_1(C_{i_1}, {Fa}_{i_2})\|_1
    \label{eq:7}
\end{equation}

where $i_1 = (0, 1)$ and $i_2 = (0, 1)$ denote the image indices. $X_{i_1}$, $C_{i_1}$, and $Fa_{i_2}$ represent the original image, irrelevant content semantics, and all forgery semantics, respectively.

\underline{Overall Loss.} The final loss function $L_{s_1}$ for stage 1 is:
\begin{equation}
    L_{s_1} = \rho_1L_{cls} + \rho_2L_{rec}
    \label{eq:8}
\end{equation}

where $\rho_1$ and $\rho_2$ are trade-off hyperparameters.

\subsection{Training Stage 2}
\textbf{Encoder2.} In $Encoder_2$, we utilize a portion of $Encoder_1$ responsible for extracting all forgery semantics, followed by convolutional layers to further disentangle these semantics into unique and common forgery semantics. The supplementary material illustrates the process for inputs $X_1$ and $Xh_1$. The overall process is represented as:
\begin{equation}
    Encoder_2(X_i, Xh_i) = Fu_i, Fc_i
    \label{eq:9}
\end{equation}

where $i = (0, 1)$ denotes the image indices. $X_i$, $Xh_i$, $Fu_i$, and $Fc_i$ represent RGB information, high-frequency features, unique forgery semantics, and common forgery semantics, respectively.
\\
\textbf{Decoder2.} In $Decoder_2$, we recognize that unique and common forgery semantics contain more local information. Hence, we design two dual-channel networks, each comprising only convolutional layers, and merge them during the process to reconstruct image semantics. We continue to use self-reconstruction and cross-reconstruction to improve decoupling capability \cite{Yan2023UCF}. The supplementary material illustrates the process for inputs $Fu_1$ and $Fc_1$ to obtain self-reconstructed image semantics $Srf_1$. The entire process is summarized as:
\begin{equation}
    Decoder_2(Fu_{i_1}, Fc_{i_2}) = 
    \begin{cases} 
      {Srf}_{i_1} & \text{if } i_1 = i_2 \\
      {Crf}_{i_1} & \text{if } i_1 \neq i_2 
    \end{cases}
    \label{eq:10}
\end{equation}

where $i_1 = (0, 1)$ and $i_2 = (0, 1)$ denote the image indices. $Fu_{i_1}$, $Fc_{i_2}$, $Srf_{i_1}$, and $Crf_{i_1}$ represent unique forgery semantics, common forgery semantics, self-reconstructed image semantics, and cross-reconstructed image semantics, respectively.
\\
\textbf{Objective Function.} The overall loss function of the framework combines four distinct components using a weighted sum: unique forgery semantics detection loss, common forgery semantics detection loss, contrastive loss, and reconstruction loss.

\underline{Classification Loss.} For detecting unique forgery semantics, we use cross-entropy loss:
\begin{equation}
    L_{cls_1} = L_{ce}(Detector_2(Fu_i), S_i)
    \label{eq:11}
\end{equation}

where $i = (0, 1)$ denotes the image indices. $L_{ce}$, $Fu_i$, and $S_i$ represent the cross-entropy loss, unique forgery semantics, and multi-class label regarding the forgery method, respectively.

For detecting common forgery semantics, we also use cross-entropy loss:
\begin{equation}
    L_{cls_2} = L_{ce}(Detector_3(Fc_i), y_i)
    \label{eq:12}
\end{equation}

where $i = (0, 1)$ denotes the image indices. $L_{ce}$, $Fc_i$, and $y_i$ (fake, real) represent the cross-entropy loss, common forgery semantics, and binary classification labels, respectively.

The total classification loss is:
\begin{equation}
    L_{cls} = \rho_3L_{cls_1} + \rho_4L_{cls_2}
    \label{eq:13}
\end{equation}

where $\rho_3$ and $\rho_4$ are trade-off hyperparameters.

\underline{Contrastive Loss.} To enhance the encoding capability of the encoder for different image semantics, we utilize contrastive loss $L_{con}$. The loss function  minimizes the distance between the anchor image semantics \(f^{a}_i\) and its corresponding positive image semantics \(f^+_i\), while simultaneously maximizing the distance between the anchor image semantics and its corresponding negative image semantics \(f^-_i\) \cite{Yan2023UCF}. Here, positive samples indicate the same source, and negative samples indicate different sources. For real images, the same source refers to other real images, and different sources refer to fake images. For fake images, the same source refers to other fake images created using the same forgery method, and different sources refer to real images. The formula is as follows:
\begin{equation}
    L_{con} = \max\{0, a+\|f^{a}_i-f^+_i\|_2-\|f^{a}_i-f^-_i\|_2\}
    \label{eq:14}
\end{equation}

where \(i = (0u,0c,1u,1c)\) represents unique forgery semantics for fake images, common forgery semantics for fake images, unique forgery semantics for real images, and common forgery semantics for real images, respectively. \(a\) is a trade-off hyperparameter.

\underline{Reconstruction Loss.} Self-reconstruction and cross-reconstruction are employed to reconstruct image semantics with reconstruction loss $L_{rec}$ that can be written as :
\begin{equation}
    L_{rec} = \|Fa_{i_1}-Decoder_2(Fc_{i_1}, Fu_{i_2})\|_1.
    \label{eq:15}
\end{equation}

where $i_1$ = (0, 1) and $i_2$ = (0, 1) denote image indices. $Fa_{i_1}$, $Fc_{i_1}$, and $Fu_{i_2}$ represent all forgery semantics, common forgery semantics, and unique forgery semantics, respectively.

\underline{Overall Loss.} The final loss function $L_{s_2}$ for the training stage 2 is:
\begin{equation}
     L_{s_2} = L_{cls} + \rho_5L_{con} + \rho_6L_{rec}
    \label{eq:16}
\end{equation}

where $\rho_5$ and $\rho_6$ are trade-off hyperparameters.

\section{Experimental Settings}
\textbf{Datasets.} We trained our model on FaceForensics++ (FF++) \cite{Rossler2019Faceforensics++} and evaluated it on FF++, DeepfakeDetection (DFD) \cite{Google2019Deepfakes}, Deepfake Detection Challenge (DFDC) \cite{2021Deepfake}, and Celeb-DF \cite{Li2020Celeb-DF}. FF++ includes images generated by five facial manipulation algorithms: DeepFakes (DF) \cite{2017DeepFakes}, Face2Face (F2F) \cite{Thies2016Face2face}, FaceSwap (FS) \cite{Marek2018FaceSwap}, NeuralTexture (NT) \cite{Thies2019Deferred}, and FaceShifter (FST) \cite{Li2019Faceshifter}.
\\
\textbf{Implementation.} We utilized PyTorch and NVIDIA RTX 3090Ti for training. Images were resized to 256×256, while all models were trained for 20 epochs with a fixed batch size of 16. We employed SGD optimizer \cite{robbins1951stochastic} with a learning rate of \( \beta = 5 \times 10^{-4} \). In training stage 1, we set \( \rho_1 = 1.0 \) and \( \rho_2 = 0.3 \) in Equation (\ref{eq:8}). For training stage 2, we set \( \rho_3 = 0.1 \) and \( \rho_4 = 1.0 \) in Equation (\ref{eq:13}), \( a = 3.0 \) in Equation (\ref{eq:14}), and \( \rho_5 = 0.05 \) and \( \rho_6 = 0.3 \) in Equation (\ref{eq:16}).
\\
\textbf{Evaluation Metrics.} We employ the Area Under Curve (AUC) metric for performance evaluation, consistent with prior studies \cite{he2016deep, tan2019efficientnet, Chollet2017Xception, Luo2021Generalizing, Qian2020f3net, Yan2023UCF, Lin2024Fairness}.

\subsection{Results}
\textbf{Intra-domain Sub-datasets Performance:} Our approach, as depicted in Table~\ref{tab:1}, effectively separates domain-specific forgery, mitigating over-fitting risks. Compared to Resnet-50 \cite{he2016deep}, EfficientNet-B4 \cite{tan2019efficientnet}, Xception \cite{Chollet2017Xception}, SRM \cite{Luo2021Generalizing}, F3-Net \cite{Qian2020f3net}, UCF \cite{Yan2023UCF}, and Lin \emph{et al.} \cite{Lin2024Fairness}, our method consistently achieves higher AUC scores across all sub-datasets.

\begin{table}[h]
\begin{center}
\begin{tabular}{|c|c|c|c|c|c|}
\hline
\multicolumn{1}{|c|}{\multirow{2}{*}{Method}} & \multicolumn{5}{c|}{AUC(\%)}\\
\cline{2-6}
\multicolumn{1}{|c|}{} & F2F \cite{Thies2016Face2face} & FS \cite{Marek2018FaceSwap} & NT \cite{Thies2019Deferred} & DF \cite{2017DeepFakes} & FST \cite{Li2019Faceshifter} \\
\hline\hline
\multirow{1}{*}{ResNet-50 \cite{he2016deep}} & 93.76 & 93.30 & 83.43 & 93.34 & 92.25 \\
\hline
\multirow{1}{*}{EfficientNet-B4 \cite{tan2019efficientnet}} & 97.41 & 97.10 & 90.87 & 97.02 & 96.28 \\
\hline
\multirow{1}{*}{Xception \cite{Chollet2017Xception}} & 96.92 & 95.85 & 94.00 & 97.47 & 95.62 \\
\hline
\multirow{1}{*}{SRM \cite{Luo2021Generalizing}} & 96.49 & 97.59 & 92.66 & 97.64 & 97.55 \\
\hline
\multirow{1}{*}{F3-Net \cite{Qian2020f3net}} & 96.56 & 94.14 & 93.15 & 97.67 & 96.80 \\
\hline
\multirow{1}{*}{UCF \cite{Yan2023UCF}} & 97.12 & 97.46 & 91.99 & 97.40 & 97.31 \\
\hline
\multirow{1}{*}{Lin \emph{et al.} \cite{Lin2024Fairness}} & 98.37 & 97.97 & 95.06 & 98.86 & 98.41 \\
\hline
\multirow{1}{*}{Ours} & \textbf{99.15} & \textbf{99.36} & \textbf{96.23} & \textbf{99.29} & \textbf{99.13} \\
\hline
\end{tabular}
\end{center}
\caption{Methods trained on FF++ are evaluated intra-domain on sub-datasets separated by five types of forgeries: F2F, FS, NT, DF, and FST. Best results are highlighted in bold.}
\label{tab:1}
\end{table}

\noindent\textbf{Cross-domain Datasets Performance:} Trained on FF++, our method outperform other benchmarks when tested on FF++ \cite{Thies2016Face2face}, Celeb-DF \cite{Marek2018FaceSwap}, DFD \cite{Thies2019Deferred}, and DFDC \cite{2017DeepFakes}, as summarized in Table~\ref{tab:2}. This demonstrates superior generalization capabilities, resulting in the most accurate detection outcomes.

\begin{table}[h]
\begin{center}
\begin{tabular}{|c|c|c|c|c|}
\hline
\multicolumn{1}{|c|}{\multirow{2}{*}{Method}} & \multicolumn{4}{c|}{AUC(\%)}\\
\cline{2-5}
\multicolumn{1}{|c|}{} & FF++ \cite{Thies2016Face2face} & Celeb-DF \cite{Marek2018FaceSwap} & DFD \cite{Thies2019Deferred} & DFDC \cite{2017DeepFakes} \\
\hline\hline
\multirow{1}{*}{ResNet-50 \cite{he2016deep}} & 91.06 & 64.78 & 72.94 & 53.38 \\
\hline
\multirow{1}{*}{EfficientNet-B4 \cite{tan2019efficientnet}} & 95.63 & 67.80 & 76.81 & 56.59 \\
\hline
\multirow{1}{*}{Xception \cite{Chollet2017Xception}} & 95.93 & 69.37 & 78.08 & 56.87 \\
\hline
\multirow{1}{*}{SRM \cite{Luo2021Generalizing}} & 96.30 & 68.08 & 77.57 & 58.22 \\
\hline
\multirow{1}{*}{F3-Net \cite{Qian2020f3net}} & 95.64 & 67.62 & 80.51 & 55.96 \\
\hline
\multirow{1}{*}{UCF \cite{Yan2023UCF}} & 96.17 & 70.48 & 75.68 & 55.20 \\
\hline
\multirow{1}{*}{Lin \emph{et al.} \cite{Lin2024Fairness}} & 97.68 & 75.19 & 80.56 & 62.18 \\
\hline
\multirow{1}{*}{Ours} & \textbf{98.58} & \textbf{76.94} & \textbf{83.02} & \textbf{62.55} \\
\hline
\end{tabular}
\end{center}
\caption{Methods trained on FF++ and tested on multiple datasets, including Celeb-DF, DFD, and DFDC, are presented. Best results are highlighted in bold.}
\label{tab:2}
\end{table}

\subsection{Ablation Study}
We evaluated the model combining RGB and high-frequency features along with its variations of multi-scale high-frequency feature extraction (MHFE) and multi-scale high-frequency feature fusion (MHFF) on the FF++ database. RGB represents using only RGB images, and High-frequency represents using only high-frequency features. All models were trained on FF++ and tested on four datasets, as shown in Table~\ref{tab:3}, demonstrating the complementary nature of the two modalities and the effectiveness of each module.

\begin{table}[h]
\begin{center}
\begin{tabular}{|c|c|c|c|c|}
\hline
\multicolumn{1}{|c|}{\multirow{2}{*}{Method}} & \multicolumn{4}{c|}{AUC(\%)}\\
\cline{2-5}
\multicolumn{1}{|c|}{} & FF++ \cite{Thies2016Face2face} & Celeb-DF \cite{Marek2018FaceSwap} & DFD \cite{Thies2019Deferred} & DFDC \cite{2017DeepFakes} \\
\hline\hline
\multirow{1}{*}{RGB} & 97.77 & 74.80 & 79.20 & 61.18 \\
\hline
\multirow{1}{*}{High-frequency} & 96.90 & 73.55 & 81.57 & 60.75 \\
\hline
\multirow{1}{*}{RGB + High-frequency(Fusion)} & 98.33 & 73.70 & 82.19 & 61.11 \\
\hline
\multirow{1}{*}{Fusion + MHFE} & 98.50 & 74.72 & 80.03 & 60.92 \\
\hline
\multirow{1}{*}{Fusion + MHFF} & 98.02 & 72.72 & 80.44 & 61.18 \\
\hline
\multirow{1}{*}{Fusion + MHFE + MHFF} & \textbf{98.58} & \textbf{76.94} & \textbf{83.02} & \textbf{62.55} \\
\hline
\end{tabular}
\end{center}
\caption{Ablation study on FF++. The best results are highlighted in bold.}
\label{tab:3}
\end{table}

\subsection{Visualization}
\begin{figure}[h]
  \centering
\includegraphics[height=6.5cm,width=\textwidth,keepaspectratio]{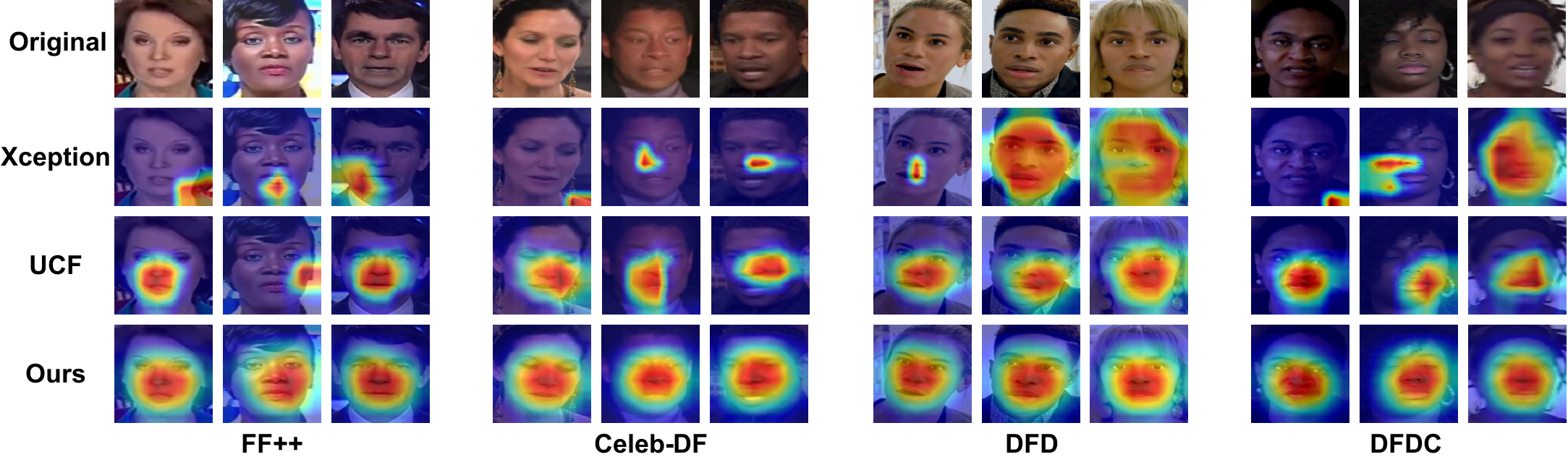}
  \caption{Visualization of Grad-CAM for Xception, UCF, and our approach across intra-domain (FF++) and cross-domain datasets (Celeb-DF, DFD, and DFDC).
  }
  \label{figure:Visualization}
\end{figure}

For a more intuitive demonstration of our method's effectiveness, we visualize the Grad-CAM \cite{selvaraju2017gradcam} of Xception, UCF, and our method, as shown in Figure~\ref{figure:Visualization}. Grad-CAM reveals that Xception tends to overfit to small local regions or focus on content noises outside facial regions when unconstrained. Although UCF performs well on certain irrelevant facial content semantics, it fails in certain cases, 
particularly when irrelevant facial semantics represent individuals with a dark skin tone, hindering its generalization performance. In contrast, our method consistently focuses on tracing the common clues in the DeepFakes left by different generating models, regardless of irrelevant facial content semantics, showcasing outstanding generalization performance.

\section{Conclusion}
In this paper, we propose a semantics decoupling approach for training a DeepFake detector, achieving satisfying forensics performance. In particular, by extracting and analyzing common forgery semantics of different DeepFake technologies, the proposed method is validated to be capable of highly generalizable DeepFake detection. Also, justified by the ablation study, the modules designed in our proposed method effectively enhance both the intra-domain and cross-domain detection performance. Our work also provides insights for discerning AI-generated content by employing semantics disentanglement. Enhancing the independence and effectiveness of forgery semantics may be crucial for obtaining high-quality forensics models.

\section*{Acknowledgement}
This work was supported in part by the National Natural Science Foundation of China under Grant 62262041, and in part by the Jiangxi Provincial Natural Science Foundation under Grant 20232BAB202011.
\bibliography{egbib}
\newpage
\begin{center}  
{\fontsize{16}{16pt}\selectfont\textbf{Decoupling Forgery Semantics for Generalizable Deepfake Detection: Supplementary Material}}  
\end{center}


\runninghead{YE ET AL.}{Decoupling Forgery Semantics}

\def\eg{\emph{e.g}\bmvaOneDot}
\def\Eg{\emph{E.g}\bmvaOneDot}
\def\etal{\emph{et al}\bmvaOneDot}



\setcounter{section}{0}
\section{The Network Details of Training Stage 1}
\begin{figure}[h]
  \centering
  \includegraphics[height=6.5cm,width=\textwidth,keepaspectratio]{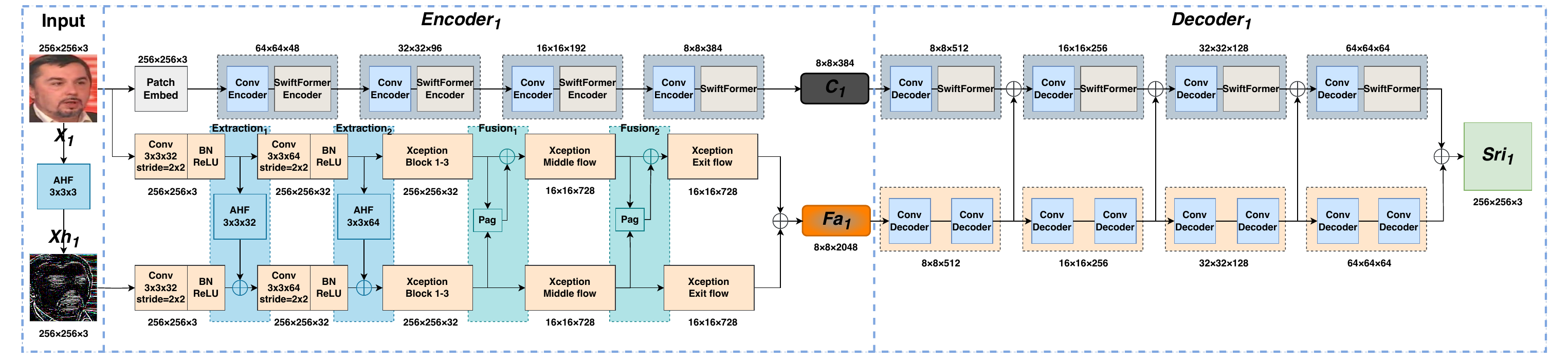}
  \caption{The architecture details of $Encoder_1$ and $Decoder_1$ in our proposed method. (\textbf{Left}) In $Encoder_1$, $Extraction_1$ and $Extraction_2$ constitute a multi-scale high-frequency feature extraction module (MHFE), while $Fusion_1$ and $Fusion_2$ form a multi-scale high-frequency feature fusion module (MHFF). (\textbf{Right}) In $Decoder_1$, convolutional layers and Swiftformer are used to reconstruct images.}
  \label{figure:1}
\end{figure}
\noindent\textbf{Encoder1.} We use SwiftFormer-L1 \cite{Shaker2023SwiftFormer} to extract irrelevant content from images and employ Xception \cite{Chollet2017Xception} for extracting RGB information and high-frequency features. Figure~\ref{figure:1} illustrates the process for inputs $X_1$ and $Xh_1$, obtaining irrelevant content semantics $C_1$ and all forgery semantics $Fa_1$. Consistently, for inputs $X_0$ and $Xh_0$, we obtain irrelevant content semantics $C_0$ and all forgery semantics $Fa_0$.
\\
\textbf{Decoder1.} In $Decoder_1$, we designed a dual-channel network. One channel simultaneously uses convolutional layers and SwiftFormer \cite{Shaker2023SwiftFormer} to process irrelevant content semantics, while the other channel solely uses convolutional layers to process all forgery semantics. Figure~\ref{figure:1} illustrates the process for inputs $C_1$ and $Fa_1$, resulting in the self-reconstructed image $Sri_1$. Similarly, inputs $C_0$ and $Fa_0$ yield the self-reconstructed image $Sri_0$. Additionally, inputs $C_1$ and $Fa_0$ produce the cross-reconstructed image $Cri_1$, while inputs $C_0$ and $Fa_1$ generate the cross-reconstructed image $Cri_0$.

\begin{figure}[h]
  \centering
  \includegraphics[height=6.5cm,width=\textwidth,keepaspectratio]{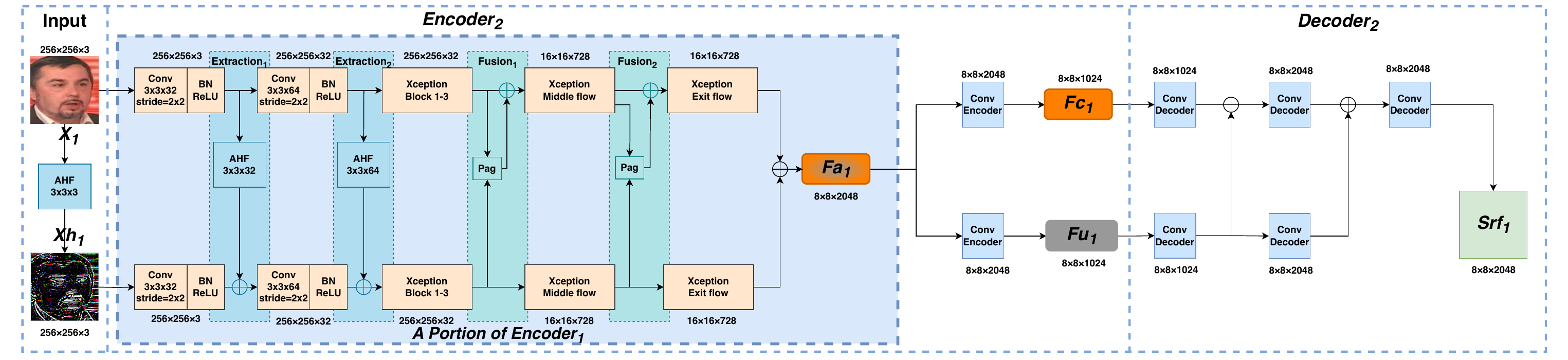}
  \caption{The architecture details of $Encoder_2$ and $Decoder_2$ in our proposed method. (\textbf{Left}) In $Encoder_2$ of training stage 2, a branch first utilizes $Encoder_1$ to extract all forgery semantics for extracting common forgery semantics. (\textbf{Right}) In $Decoder_2$, both branches solely employ convolutional layers to reconstruct forgery semantics.}
  \label{figure:2}
\end{figure}
\setcounter{section}{1}
\section{The Network Details of Training Stage 2}

\noindent\textbf{Encoder2.} In $Encoder_2$, we utilize a portion of $Encoder_1$ responsible for extracting all forgery semantics to extract all forgery semantics, and then employ convolutional layers further disentangle these semantics into unique and common forgery semantics. Figure~\ref{figure:2} illustrates the process for inputs $X_1$ and $Xh_1$, obtaining unique forgery semantics $Fu_1$ and common forgery semantics $Fc_1$. Similarly, for inputs $X_0$ and $Xh_0$, we obtain unique forgery semantics $Fu_0$ and common forgery semantics $Fc_0$.
\\
\textbf{Decoder2.} In $Decoder_2$, we designed two dual-channel networks, each comprising only convolutional decoders, and merged them during the process to reconstruct image semantics. Figure~\ref{figure:2} illustrates the process for input $Fc_1$ and $Fu_1$ to obtain self-reconstructed image semantics $Srf_1$. Consistently, for inputs $Fc_0$ and $Fu_0$, we obtain self-reconstructed image semantics $Srf_0$. Additionally, inputs $Fc_1$ and $Fu_0$ yield cross-reconstructed image semantics $Crf_1$, while inputs $Fc_0$ and $Fu_1$ yield cross-reconstructed image semantics $Crf_0$.


\end{document}